\documentclass{article}
\usepackage{graphicx,mlspconf}
\usepackage[cmex10]{amsmath}
\usepackage{amsfonts}
\usepackage{amssymb}

\usepackage[caption = false,font = footnotesize]{subfig}
\usepackage{booktabs}
\usepackage{url}

\toappear{Accepted for presentation at 2018 IEEE International Workshop on Machine Learning for Signal Processing (MLSP)}

\newcommand{\vect}[1]{\mathbf{#1}}

\title{RECURRENT NEURAL NETWORKS WITH FLEXIBLE GATES \\ USING KERNEL ACTIVATION FUNCTIONS}

\makeatletter
\def\@name{ \emph{Simone~Scardapane}$^*$$^\dagger$, \emph{Steven~Van~Vaerenbergh}$^\ddagger$, \emph{Danilo~Comminiello}$^\dagger$,\\ \emph{Simone~Totaro}$^\dagger$, and~\emph{Aurelio~Uncini}$^\dagger$\thanks{$^*$ Corresponding author e-mail: simone.scardapane@uniroma1.it.}\vspace{1em}}
\makeatother

\address{%
    \tabular{c}
		$^\dagger$ Sapienza University of Rome, Italy
	\endtabular
	\hskip 0.5in
    \tabular{c}
		$^\ddagger$ University of Cantabria, Spain
	\endtabular
}

\begin{document}
%\ninept
%

\maketitle
\begin{abstract}
Gated recurrent neural networks have achieved remarkable results in the analysis of sequential data. Inside these networks, gates are used to control the flow of information, allowing to model even very long-term dependencies in the data. In this paper, we investigate whether the original gate equation (a linear projection followed by an element-wise sigmoid) can be improved. In particular, we design a more flexible architecture, with a small number of adaptable parameters, which is able to model a wider range of gating functions than the classical one. To this end, we replace the sigmoid function in the standard gate with a non-parametric formulation extending the recently proposed kernel activation function (KAF), with the addition of a residual skip-connection. A set of experiments on sequential variants of the MNIST dataset shows that the adoption of this novel gate allows to improve accuracy with a negligible cost in terms of computational power and with a large speed-up in the number of training iterations.
\end{abstract}
\section{Introduction}
\label{sec:introduction}

Recurrent neural networks (RNNs) have recently gained a large popularity in the analysis of sequential data, following more widespread success in the field of deep learning \cite{goodfellow2016deep}. Among all possible RNNs, gated architectures (originating from the seminal work in \cite{hochreiter1997long}) have shown to be particularly suitable at handling long, or very long, temporal dependencies in the data. While the original long short-term memory (LSTM) network dates back to twenty years ago, recent advances in computational power have allowed to scale them to multi-layered and sequence-to-sequence configurations \cite{sutskever2014sequence}, achieving significant breakthroughs in multiple fields, e.g., neural machine translation \cite{cho2014learning}.

Fundamentally, a gate is a multiplicative layer that learns to perform a `soft selection' of some content (e.g., the hidden state of the RNN), allowing the gradient and the information to flow more easily through multiple time-steps while avoiding vanishing or exploding gradients. Despite their importance, however, the role and the use of gates inside RNNs remain open questions for research. The original LSTM network was designed with two gates to have a unitary derivative \cite{hochreiter1997long}, which were later increased to three with the inclusion of a forget gate \cite{gers2000recurrent}. Subsequent research has experimented with a wide range of different configurations, including the gated recurrent unit (GRU) \cite{cho2014learning}, merging two gates into a single update gate, or even simpler architectures having a single gate, such as the minimally gated unit \cite{zhou2016minimal}, or the JANET model \cite{van2018unreasonable} (see also \cite{greff2017lstm} for a large comparison of feasible variations). From a theoretical perspective, \cite{tallec2018can} has recently shown that gates naturally arise if we assume (axiomatically) a (quasi-)invariance to time transformations of the input data.

Note that, even considering this wide range of alternative formulations (mostly in terms of how many gates are needed for an optimal architecture), the basic design of a \textit{single} gate has remained more or less constant, i.e., each gate is obtained by applying an element-wise sigmoid nonlinearity to a linear projection of the inputs and/or hidden states. Only a handful of works have explored alternative designs for this component, such as the inclusion of hidden layers \cite{gao2016deep}, or skip-connections through the gates of different layers \cite{yao2015depth}. Motivated by the possibility of improving the performance of the RNNs, in this paper we propose an extended gate architecture, which is endowed with a larger expressiveness than the standard formulation. At the same time, we try to keep the computational overhead as small as possible. To this end, we focus on replacing the sigmoid operation, extending it with a non-parametric form that is adapted independently for each cell (and for each gate) inside the gated RNN.

Our starting point is noting that a lot of work has been done in the deep learning literature for designing flexible activation functions, that could replace standard hyperbolic tangents or rectified linear units (ReLUs). These include simple parametric schemes like the parametric ReLU \cite{he2015delving}, or more elaborate formulations where the flexibility of the functions can be determined as a hyper-parameter. The latter case include maxout networks \cite{goodfellow2013maxout}, adaptive piecewise linear units \cite{agostinelli2014learning}, and kernel activation functions (KAFs) \cite{scardapane2017kafnets}. There is a good consensus in that endowing the functions with this flexibility can enhance the performance of the network, possibly allowing to simplify the architecture of the neural network itself significantly \cite{agostinelli2014learning}. However, the sigmoid function used inside a gate is different from a standard activation function, in that its behavior cannot be unrestricted (e.g., by taking negative values). Due to this, none of these proposals can be applied straightforwardly to the case of gates inside RNNs: for example, all functions based on rectifiers (such as the APL \cite{agostinelli2014learning}) are unbounded over their domain \cite{agostinelli2014learning}.

To this end, in this paper we propose an extension of the basic KAF model. A KAF is a non-parametric activation function defined in terms of a kernel expansion over a fixed dictionary \cite{scardapane2017kafnets}. Here, we combine it with a bounded nonlinearity and a residual connection to make its behavior consistent with that of a gating function (more details in Section \ref{sec:proposed_gate}). As a result, our proposed flexible gate mimics exactly a standard sigmoid at the beginning of the optimization process, but thanks to the addition of a small number of adaptable parameters, it can adapt itself based on the training data to a much larger family of shapes (see Fig. \ref{fig:kafgates_samples} for some examples).

We evaluate the proposed model on a set of standard benchmarks involving sequential formulations of the MNIST dataset (e.g., where each image is processed pixel-by-pixel). We show that a gated RNN with our proposed flexible gate can achieve higher accuracy with a faster rate of convergence, while at the same time having a small computational overhead with respect to a standard formulation.

\subsubsection*{Paper outline}
The rest of the paper is organized as follows. In Section \ref{sec:gated_rnns} we introduce the GRU model (as a representative example of gated RNN). Next, the proposed gate with flexible sigmoids is described in Section \ref{sec:proposed_gate}. We evaluate the proposal in Section \ref{sec:experimental_results}, before concluding in Section \ref{sec:conclusions}.

\section{Gated recurrent neural networks}
\label{sec:gated_rnns}

\subsection{Update equations}

Consider a generic sequential task, where at each time step $t$ we receive a new input $\vect{x}_t \in \mathbb{R}^d$. The evolution of a generic RNN can be described by the following equation:

\begin{equation}
\vect{h}_t = \phi \left( \vect{x}_t, \vect{h}_{t-1} \,;\, \boldsymbol{\theta} \right) \,,
\end{equation}

\noindent where $\vect{h}_t$ represents the internal state of the RNN, $\boldsymbol{\theta}$ is the set of adaptable parameters, and $\phi(\cdot)$ is a generic update rule. Gated RNNs implement $\phi(\cdot)$ with the presence of one or more gating functions, which control the flow of information between time steps. As we stated in Section \ref{sec:introduction}, different types of gated RNNs, with different number of gates, exist in the literature. For brevity, in the rest of the paper we will focus on the case of GRUs \cite{cho2014learning}, although our method extends immediately to LSTMs and any other gated network described in the previous section. However, GRUs represent a good compromise between accuracy and number of gates (two compared to three as in the LSTM), which is why we choose it here.

A GRU cell updates its internal state $\vect{h}_{t-1}$ as follows:

\begin{align}
\vect{u}_t & = \sigma \left( \vect{W}_u \vect{x}_t + \vect{V}_u \vect{h}_{t-1} + \vect{b}_u \right) \,, \label{eq:update_gate}\\
\vect{r}_t & = \sigma \left( \vect{W}_r \vect{x}_t + \vect{V}_r \vect{h}_{t-1} + \vect{b}_r \right) \,, \label{eq:forget_gate}\\
\vect{h}_t & = \left( 1 - \vect{u}_t \right) \circ \vect{h}_{t-1} \,\,+ \nonumber \\ 
&  \vect{u}_t \circ \text{tanh}\left( \vect{W}_h \vect{x}_t + \vect{U}_t \left( \vect{r}_t \circ \vect{h}_{t-1} \right) + \vect{b}_h \right) \label{eq:gru_final_update} \,,
\end{align} 

\noindent where \eqref{eq:update_gate} and \eqref{eq:forget_gate} are, respectively, the update gate and reset gate, $\circ$ is the element-wise multiplication, $\sigma(\cdot)$ is the standard sigmoid function, and the cell has $9$ adaptable matrices given by $\boldsymbol{\theta} = \left\{ \vect{W}_u, \vect{W}_f, \vect{W}_h, \vect{V}_u, \vect{V}_f, \vect{V}_h, \vect{b}_u, \vect{b}_f, \vect{b}_h \right\}$. Note that while the $\tanh(\cdot)$ function in \eqref{eq:gru_final_update} can be changed freely (e.g., to a ReLU function), the sigmoid function in the two gates is essential for having a correct behavior, i.e., the update vector $\vect{u}_t$ and reset vector $\vect{r}_t$ should always remain bounded in $\left[0, 1\right]$.

\subsection{Training the network}

GRUs can be used for a variety of tasks by properly manipulating the sequence of their internal states $\vect{h}_1, \vect{h}_2, \ldots$. Since in our experiments we consider the problem of classifying each sequence of data, we briefly describe here the details of the optimization approach. We underline, however, that the method we propose in the next section is agnostic to the actual task, as it acts on the basic GRU formulation.

Suppose to have $N$ different sequences $\left\{\vect{x}_t^i\right\}_{i=1}^N$, and for each of them a single class label $y^i = 1, \ldots, C$. Denote by $\vect{h}^i$ the internal state of the GRU after processing the $i$-th sequence. To obtain a classification, this is fed through another layer with a softmax activation function:

\begin{equation}
\widehat{\vect{y}}^i = \text{softmax}\left( \vect{A}\vect{h}^i + \vect{b} \right) \,,
\label{eq:softmax_layer}
\end{equation}

\noindent with $\widehat{\vect{y}}^i$ having values over the $C$-dimensional simplex. The network is trained by minimizing the average cross-entropy between the real classes and the predicted classes:

\begin{equation}
J(\boldsymbol{\theta}) = - \frac{1}{N} \sum_{i=1}^N \sum_{c=1}^C \left[ y^i = c \right] \log\left( \widehat{y}^i_j \right) \,,
\label{eq:cross_entropy}
\end{equation}

\begin{figure*}
\subfloat[$\gamma$ = 1.0]{
\includegraphics[width=0.6\columnwidth,keepaspectratio]{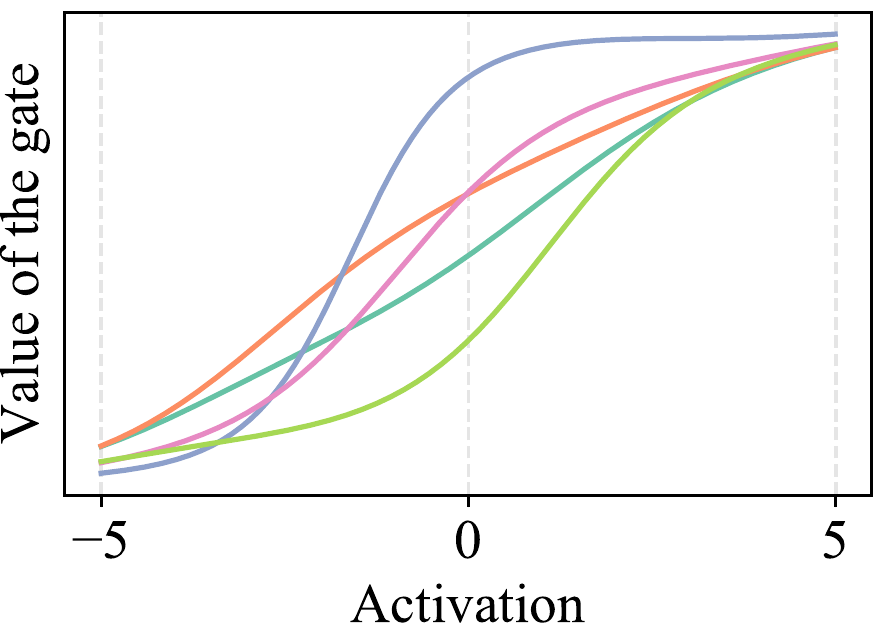}
\label{fig:kafgates_samples_small_gamma}
} \hfill
\subfloat[$\gamma$ = 0.5]{
\includegraphics[width=0.6\columnwidth,keepaspectratio]{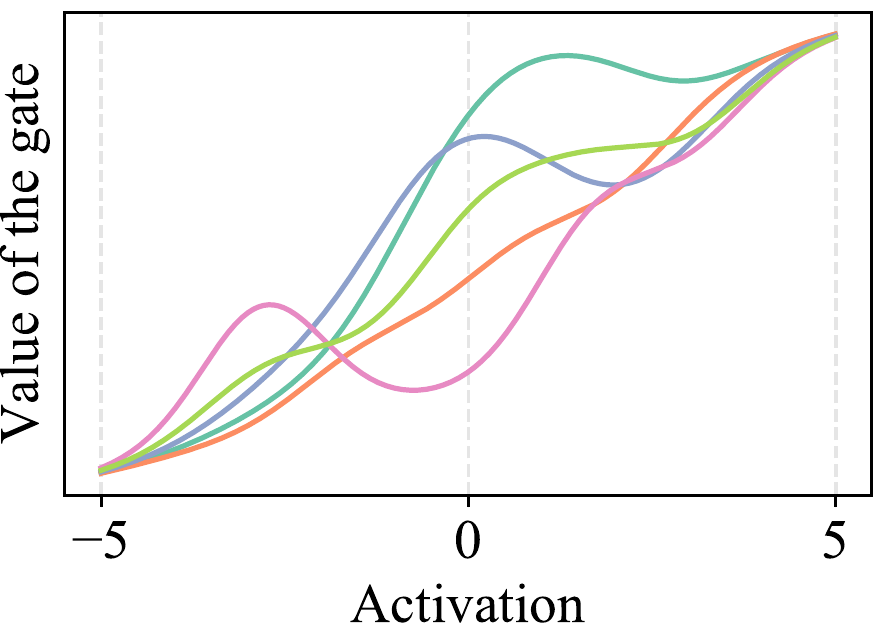}
\label{fig:kafgates_samples_medium_gamma}
} \hfill
\subfloat[$\gamma$ = 0.1]{
\includegraphics[width=0.6\columnwidth,keepaspectratio]{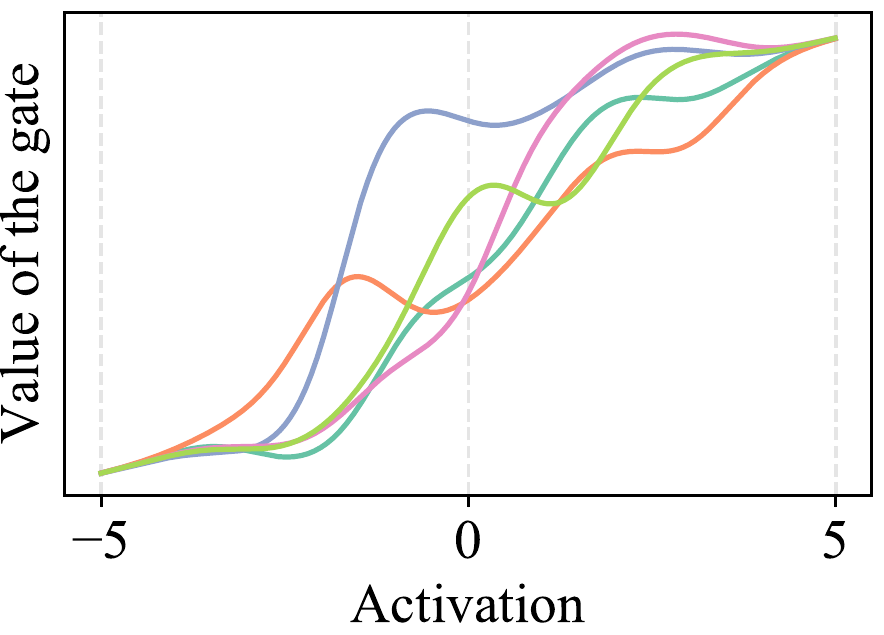}
\label{fig:kafgates_samples_large_gamma}
} \hfill
\caption{Random samples of the proposed flexible gates with different bandwidths. In all cases we sample uniformly $10$ points on the $x$-axis, while the mixing coefficients are sampled from a normal distribution. $y$-axis always goes from $0$ to $1$.}
\label{fig:kafgates_samples}
\end{figure*}

\noindent where $\widehat{y}^i_j$ is the $j$-th element of the prediction vector $\widehat{\vect{y}}^i$, and $\left[\cdot\right]$ is $1$ if its argument is true, $0$ otherwise. Minimization of \eqref{eq:cross_entropy} is obtained by unrolling the network over all time steps via back-propagation through time (BPTT) \cite{goodfellow2016deep}. While this covers the basic mathematical framework, in practice several methods can be used to stabilize and improve the convergence of BPTT, including gradients' clipping \cite{sutskever2014sequence}, multiple variations of dropout \cite{gal2016theoretically}, or regularizing appropriately weights and/or changes in activations during training \cite{krueger2016regularizing}. 

\section{Proposed gate with flexible sigmoid}
\label{sec:proposed_gate}

What we propose in this paper is to replace the sigmoid in \eqref{eq:update_gate} and \eqref{eq:forget_gate} with another (scalar) function with higher flexibility, while (a) keeping the overall `sigmoid-like' behavior, and (b) maintaining a low computational overhead. Our proposal builds upon the KAF \cite{scardapane2017kafnets}, which was originally designed as a replacement for standard activation functions, e.g., the ReLU. In this section we briefly describe the KAF formulation before introducing our extension. 

\subsection{Kernel activation functions}

A KAF is defined as a one-dimensional kernel expansion:

\begin{equation}
\text{KAF}(s) = \sum_{i=1}^D \alpha_i \kappa\left(s, d_i\right) \,,
\label{eq:kaf}
\end{equation}

\noindent where $s$ is a generic input to the activation function, $\kappa(\cdot, \cdot): \mathbb{R} \times \mathbb{R} \rightarrow \mathbb{R}$ is a valid kernel function, $\left\{\alpha_i\right\}_{i=1}^D$ are called mixing coefficients, and $\left\{d_i\right\}_{i=1}^D$ are called the dictionary elements. To make back-propagation tractable, differently from a standard kernel method the $D$ elements of the dictionary are fixed beforehand and shared across the entire network. In particular, we let $D$ as a user-chosen hyper-parameter, and we sample $D$ values over the $x$-axis, uniformly around zero. Basically, higher values of D will correspond to an increased flexibility of the function, together with an increase in the number of free parameters. Mixing coefficients are adapted through standard back-propagation, independently for every neuron, which can be realized efficiently through vectorized operations \cite{scardapane2017kafnets}.

The kernel function $\kappa(\cdot, \cdot)$ only needs to respect the positive semi-definiteness property, and for our experiments we use the 1D Gaussian kernel defined as:
\begin{equation}
\kappa(s, d_i) = \exp\left\{-\gamma\left(s - d_i\right)^2\right\} \,,
\label{eq:gaussian_kernel}
\end{equation}
where $\gamma \in \mathbb{R}$ is a kernel parameter, i.e., the inverse bandwidth. The parameter $\gamma > 0$ defines the range of influence of each $\alpha_i$ element. For selecting it, we adopt the rule-of-thumb proposed in \cite{scardapane2017kafnets}:

\begin{equation}
\gamma = \frac{1}{6\Delta^2} \,,
\label{eq:sigma_rule_of_thumb}
\end{equation}

\noindent where $\Delta$ is the resolution of the dictionary elements. Additionally, we have found beneficial to let $\gamma$ adapt independently for each KAF, always through back-propagation.

\subsection{Flexible gating functions using KAFs}

We cannot use \eqref{eq:kaf} straightforwardly because (a) it is unbounded, and (b) using the Gaussian kernel, it goes to $0$ for $s \rightarrow \pm \infty$ in both directions. We propose to alleviate these problems by using the following formulation for the flexible gate:

\begin{equation}
\sigma_{\text{KAF}}(s) = \sigma \left( \frac{1}{2}\text{KAF}(s) + \frac{1}{2}s \right) \,,
\label{eq:proposed_flexible_gate}
\end{equation}

\noindent where the sigmoid $\sigma$ on the right-hand side keeps the boundedness of the function, while the addition of the residual term $s$ ensures that \eqref{eq:proposed_flexible_gate} behaves as a standard sigmoid outside the range of the dictionary. In Fig. \ref{fig:kafgates_samples} we show some realizations of \eqref{eq:proposed_flexible_gate} for different choices of the mixing coefficients and $\gamma$. It can be seen that the functions can represent a wide array of different shapes, all consistent with the general behavior of a gating function.

\begin{figure}
\includegraphics[width=0.95\columnwidth,keepaspectratio]{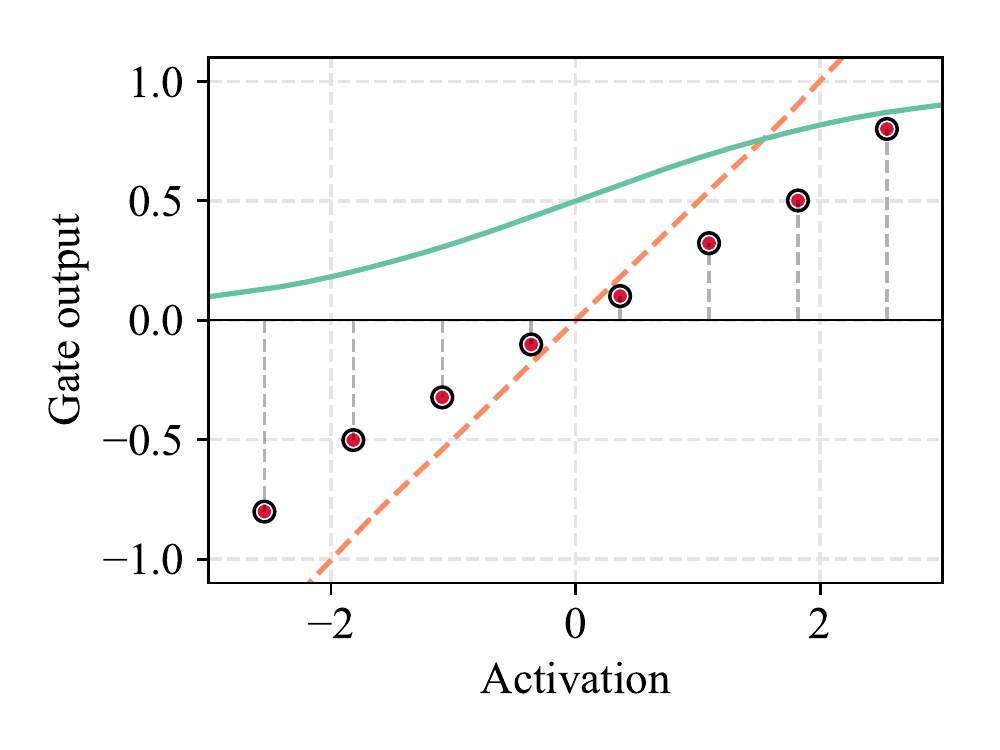}
\caption{Example of the proposed KAF gate when initialized as a standard sigmoid. The dashed line is $\text{KAF}(s)$ in \eqref{eq:proposed_flexible_gate}; markers show its mixing coefficients; the solid green line is the final output of the gate.}
\label{fig:kafgate_initialization_example}
\end{figure}

In fact, to simplify optimization we also initialize the mixing coefficients to approximate the identity function, so that the flexible gate behaves as a sigmoid in the initial stage of training. In order to do this, we apply kernel ridge regression on the dictionary to select the initial values for the mixing coefficients:

\begin{equation}
\boldsymbol{\alpha} = \left(\vect{K} + \varepsilon\vect{I}\right)^{-1}\vect{d} \,,
\label{eq:alpha_initialization}
\end{equation}

\noindent where $\boldsymbol{\alpha}$ is the vector of mixing coefficients, $\vect{d}$ is the vector of dictionary elements, $\vect{I}$ is the identity matrix of appropriate size, and $\varepsilon > 0$ is a small scalar coefficient to ensure stability of the matrix inversion (we use $\varepsilon=10^{-4}$ in the experiments). By using this initialization, all gates will behave identically to a standard sigmoid at the beginning (an example of initialization is shown in Fig. \ref{fig:kafgate_initialization_example}). In our proposed GRU, we then use a different set of mixing coefficients for each forget gate and update gate.

\section{Experimental results}
\label{sec:experimental_results}

\subsection{Experimental setup}

To evaluate the proposed algorithm, we compare a standard GRU with a GRU endowed with flexible gates as in \eqref{eq:proposed_flexible_gate}. We use a standard set of sequential benchmarks constructed from the MNIST\footnote{\url{http://yann.lecun.com/exdb/mnist/}} dataset, which are commonly used for testing long-term dependencies in gated RNNs \cite{zhang2016architectural}. MNIST is an image classification dataset composed of 60000 images for training (and 10000 for testing), each belonging to one out of ten classes. Each image is of dimension $28 \times 28$ with black-and-white pixels. From this, we construct three sequential problems:

\begin{description}
\item[\textbf{Row-wise MNIST} (R-MNIST)] Each image is processed sequentially, row-by-row, i.e., we have sequences of length $28$, each represented by the value of $28$ pixels.
\item[\textbf{Pixel-wise MNIST} (P-MNIST)] Each image is represented as a sequence of $784$ pixels, read from left to right and from top to bottom from the original image.
\item[\textbf{Permuted P-MNIST} (PP-MNIST)] Similar to P-MNIST, but the order of the pixels is shuffled using a (fixed) permutation matrix.
\end{description}

\noindent P-MNIST and PP-MNIST are particularly challenging because of the need of processing relatively long-term dependencies in the data. 

GRUs have an internal state of dimensionality $100$, and we include an additional batch normalization step \cite{goodfellow2016deep} before the output layer in \eqref{eq:softmax_layer} to stabilize training in the presence of long sequences. We train using the Adam optimization algorithm with BPTT on mini-batches of $32$ elements, with an initial learning rate of $0.001$, and we clip all gradients updates (in norm) to $1.0$. For the proposed gating function, we initialize the dictionary from $10$ elements equispaced in $\left[ -4.0, 4.0 \right]$.

Early stopping is used to decide when to finish the optimization procedure. We keep the last 10000 elements of the training set as a validation part, and we compute the average accuracy of the model every $25$ iterations, stopping whenever accuracy is not improving for at least $500$ iterations. All the code is written in PyTorch and it is run using a Tesla K80 GPU on the Google Colaboratory platform.
\subsection{Discussion of the results}

\begin{table}
{\centering\hfill{}
\setlength{\tabcolsep}{4pt}
\renewcommand{\arraystretch}{1.6}
\begin{normalsize}
\begin{tabular}{lcc}   %@{}p{0.3\columnwidth}p{0.2\columnwidth}p{0.2\columnwidth}@{}
\toprule
\textbf{Dataset} & \textbf{GRU (Standard)} & \textbf{GRU (proposed)} \\
\midrule
R-MNIST & $98.29 \pm 0.01$ & $\vect{98.67 \pm 0.02}$ \\
P-MNIST & $89.50 \pm 5.64$ & $\vect{97.34 \pm 0.61}$ \\
PP-MNIST & $86.41 \pm 6.71$ & $\vect{96.10 \pm 0.93}$ \\
\bottomrule
\end{tabular}
\end{normalsize}
}
\hfill{}
\caption{Average test accuracy obtained by a standard GRU compared with a GRU endowed with the proposed flexible gates (standard deviation is shown in brackets).}
\label{tab:results}
\end{table}

\begin{figure*}
\centering
\subfloat[Training loss]{
\includegraphics[width=0.9\columnwidth,keepaspectratio]{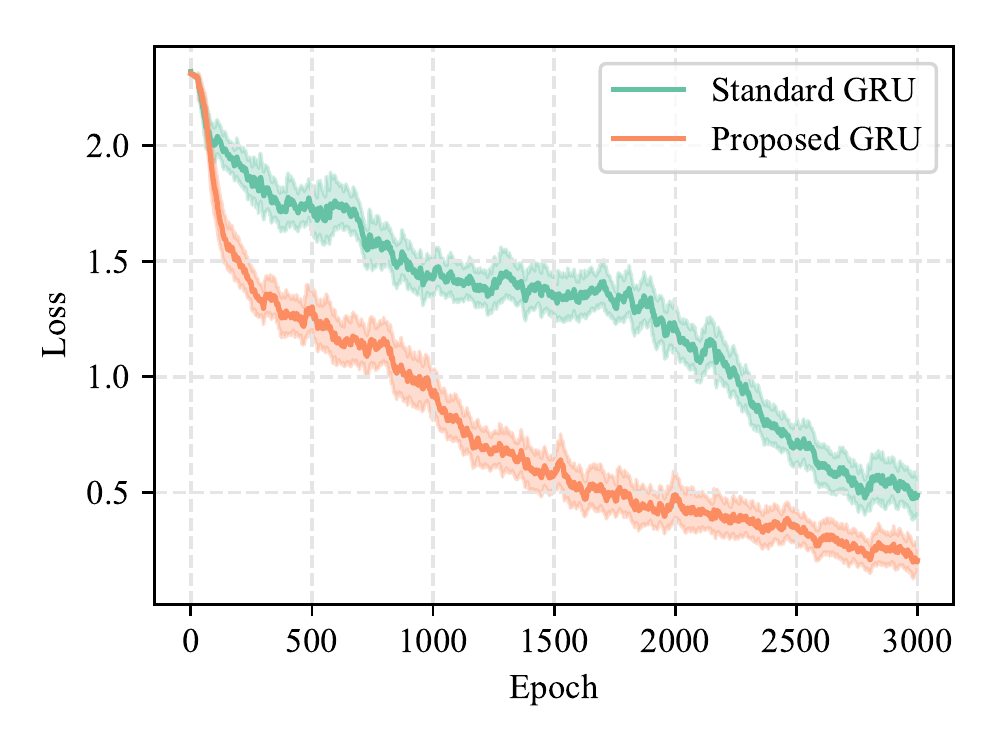}
\label{fig:convergence_pmnist_loss}
} \hfil
\subfloat[Validation accuracy]{
\includegraphics[width=0.9\columnwidth,keepaspectratio]{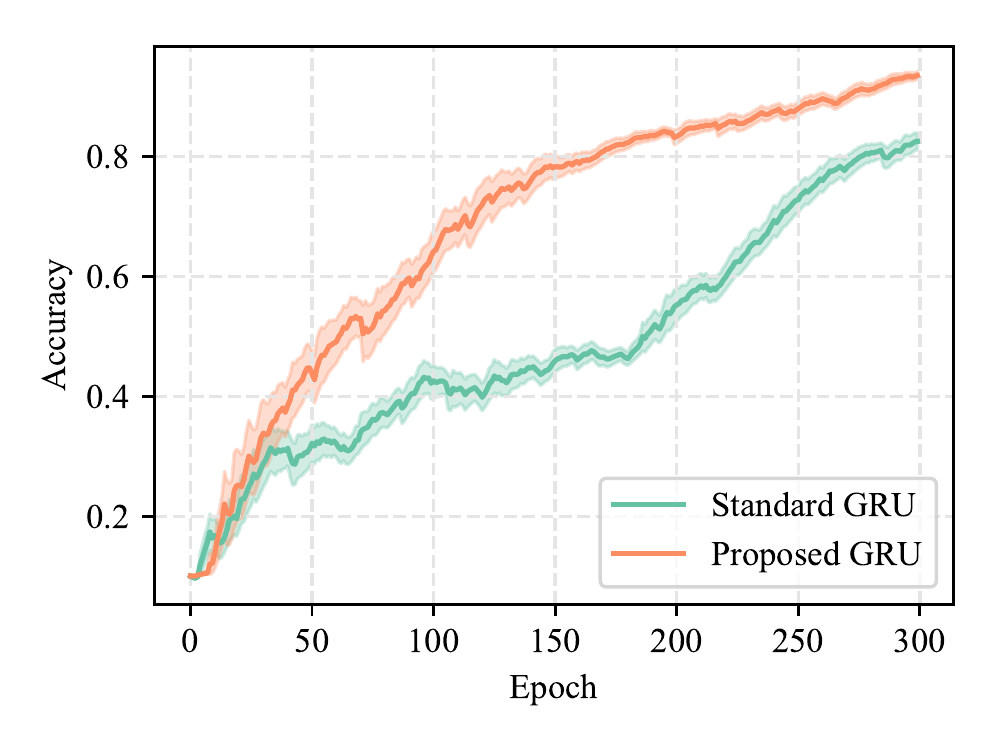}
\label{fig:convergence_pmnist_accuracy}
}
\caption{Convergence results on the P-MNIST dataset for a standard GRU and the proposed GRU. (a) Loss evolution on the training dataset (per iteration); (b) Validation accuracy (per epoch). The plots are focused on the first half of training. Shaded areas represent the variance.}
\label{fig:convergence_pmnist}
\end{figure*}

The results of the experiments averaged over $10$ different runs are given in Tab. \ref{tab:results}. We can see that the proposed GRU achieves higher test accuracy in all three cases (third column in Tab. \ref{tab:results}). Interestingly, this difference is particularly significant for datasets with long temporal dependencies (P-MNIST and PP-MNIST). Here, the standard GRU also exhibits high standard deviations due to it converging to poorer minima in some cases. The proposed GRU is able to achieve high accuracy consistently over all runs.

We conjecture that this last result is also due to the higher flexibility allowed to the optimization procedure during training. To test this, we visualize in Fig. \ref{fig:convergence_pmnist} the average loss and validation accuracy on the P-MNIST dataset of the two algorithms. We can see that the proposed GRU converges faster and more steadily, especially in the first half of training. This is consistent with the behavior found when using KAFs as activation functions, e.g., \cite{scardapane2017kafnets}.

\begin{figure}[h]
\includegraphics[width=0.9\columnwidth,keepaspectratio]{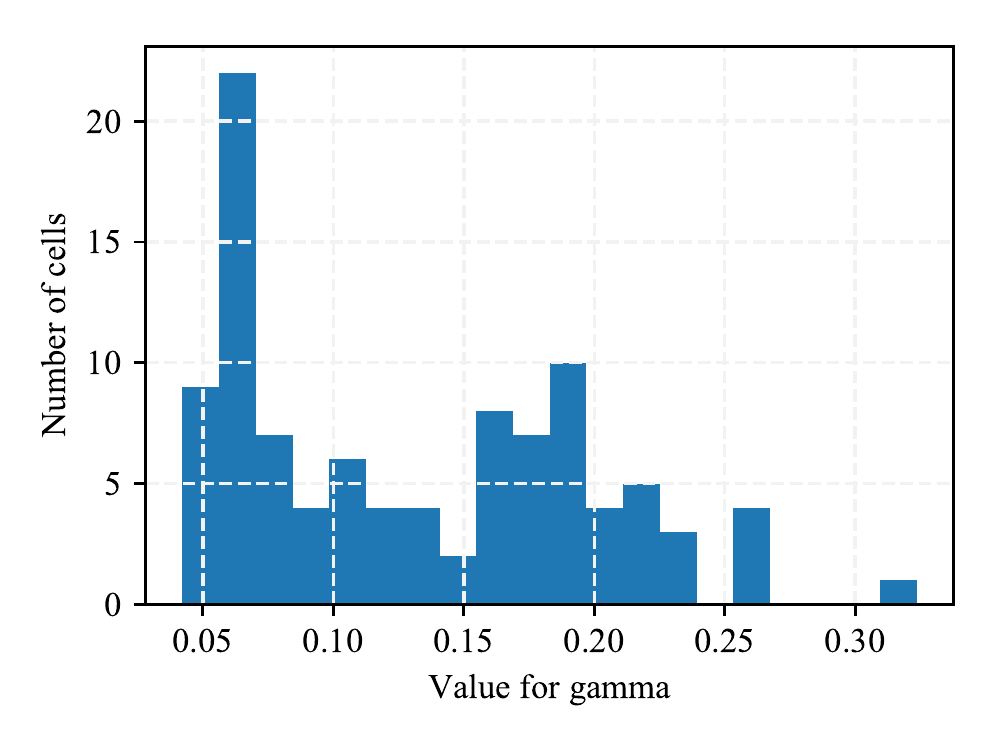}
\caption{Sample histogram of the values for $\gamma$ in \eqref{eq:gaussian_kernel}, after training, for the reset gate of the GRU.}
\label{fig:gamma_histogram}
\end{figure}

Fig \ref{fig:gamma_histogram} shows an histogram of the values of $\gamma$ in \eqref{eq:gaussian_kernel} after training the proposed GRU on the P-MNIST dataset. Interestingly, the optimal architectures can benefit from a wide range of different bandwidths for the kernel. Looking back at Fig. \ref{fig:kafgates_samples}, this translates in functions going from almost-linear  to highly nonlinear behaviors.

\subsection{Ablation study}

To conclude our experimental section, we also perform a simple ablation study on the R-MNIST dataset, by training our proposed GRU with two modifications:

\begin{itemize}
\item \textbf{Rand}: we initialize the mixing coefficients randomly instead of following the identity initialization as in \eqref{eq:alpha_initialization}.
\item \textbf{No-Residual}: we remove the residual connection from \eqref{eq:proposed_flexible_gate}, leaving only $\sigma_{\text{KAF}}(s) = \sigma\left( \text{KAF}(s) \right)$.
\end{itemize}

\begin{figure}
\includegraphics[width=0.9\columnwidth,keepaspectratio]{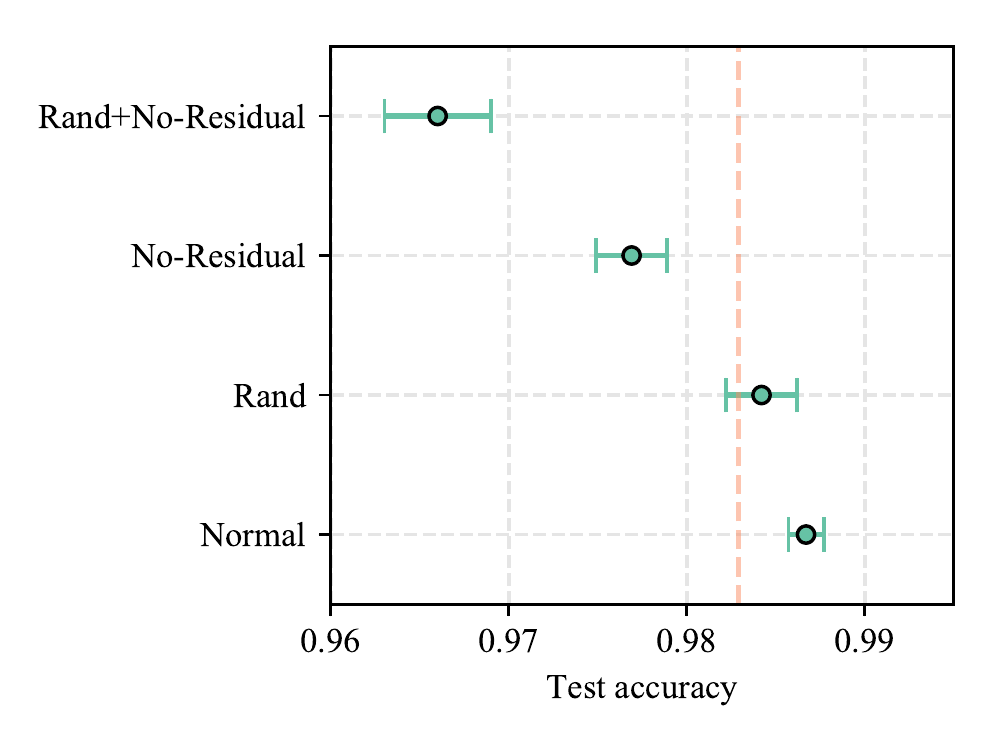}
\caption{Average results (in terms of test accuracy) of an ablation study on the R-MNIST dataset. \textbf{Rand}: we initialize the mixing coefficients randomly. \textbf{No-Residual}: we remove the residual connection in \eqref{eq:proposed_flexible_gate}. With a dashed red line we show the performance of a standard GRU.}
\label{fig:ablation_study}
\end{figure}

Results of this set of experiments are provided in Fig. \ref{fig:ablation_study}, where we also show with a red line the average test accuracy obtained by the standard GRU. We can see that removing the residual connection vastly degrades the performance, possibly because the resulting gating functions will revert to zero at their boundaries. Initializing the coefficients as the identity helps improving the accuracy by a lower margin, and also stabilizes it by reducing the variation of the results.

\section{Conclusions}
\label{sec:conclusions}

In this paper, we proposed an extension of the standard gating component used in most gated RNNs, e.g., LSTMs and GRUs. Specifically, we replace the element-wise sigmoid operation with a per-cell function endowed with a small number of parameters, that can adapt to the training data. To this end, we extend the kernel activation function in order to make its shape always consistent with a sigmoid-like behavior. The resulting function can be implemented easily in most deep learning frameworks, has a smooth behavior over its entire domain, and it imposes only a small computational overhead on the architecture. Experiments on a set of standard sequential problems with GRUs show that the proposed architecture achieve superior results (in terms of test accuracy), while at the same time converging faster (and  reliably) in terms of number of iterations due to its increased flexibility.

Future research directions involve experimenting with other gated RNNs (possibly with different numbers of gates, layers, etc.), applications, and interpreting the resulting functions with respect to the task at hand. More in general, sigmoid-like functions are essential for many other deep learning components beside gated RNNs, including softmax functions for classification, attention-based architectures, and neural memories \cite{graves2016hybrid}. An interesting question is whether our extended formulation can benefit (both in terms of accuracy and speed) these other architectures.

%\section{REFERENCES}
%\label{sec:ref}

\bibliographystyle{IEEEbib}
\bibliography{refs}

\end{document}